\pdfoutput=1

\documentclass[11pt]{article}

\usepackage{naacl2021}

\usepackage{times}
\usepackage{latexsym}

\usepackage[T1]{fontenc}

\usepackage[utf8]{inputenc}

\usepackage{microtype}

\usepackage{graphicx}
\usepackage{subfigure}
\usepackage{mathtools}
\usepackage{amsfonts}
\usepackage{booktabs}
\usepackage{multirow}
\usepackage{xspace}
\usepackage{todonotes}
\usepackage{paralist}

\newcommand{\method}{ImagiT\xspace}

\title{Generative Imagination Elevates Machine Translation}

\author{Quanyu Long$^1$, Mingxuan Wang$^2$, and Lei Li$^2$ \\
  $^1$Nanyang Technological University, Singapore \\
  $^2$ByteDance AI Lab, China\\
  \texttt{quanyu001@e.ntu.edu.sg}; \\
  \texttt{\{wangmingxuan.89, lileilab\}@bytedance.com}
  }

\begin{document}
\maketitle

\begin{abstract}
There are common semantics shared across text and images. 
Given a sentence in a source language, whether depicting the visual scene helps translation into a target language?
Existing multimodal neural machine translation methods (MNMT) require triplets of bilingual sentence - image for training and tuples of source sentence - image for inference.
In this paper, we propose \method, a novel machine translation method via visual imagination. 
\method first learns to generate visual representation from the source sentence, and then utilizes both source sentence and the ``imagined representation'' to produce a target translation.
Unlike previous methods, it only needs the source sentence at the inference time. 
Experiments demonstrate that \method benefits from visual imagination and significantly outperforms the text-only neural machine translation baselines. 
Further analysis reveals that the imagination process in \method  helps fill in missing information when performing the degradation strategy.

\end{abstract}

\section{Introduction}
\label{sec:intro}
\begin{figure}[!t]
\centering
\subfigure[Multimodal NMT]
{
\centering
\includegraphics[width=0.7\columnwidth]{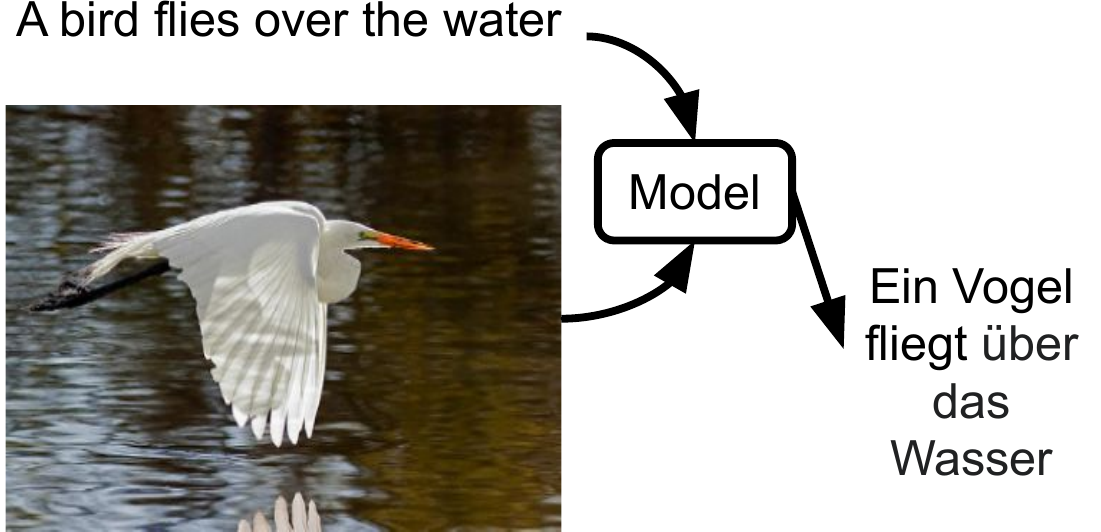}
}
\subfigure[\method]
{
\centering
\includegraphics[width=0.95\columnwidth]{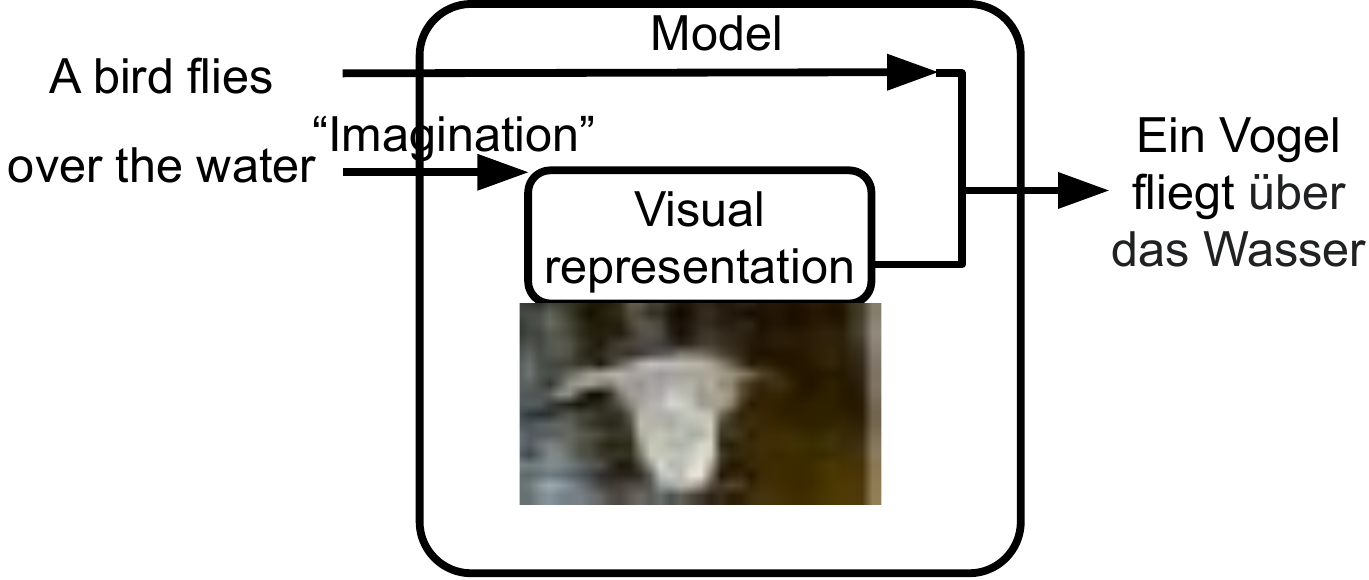}
}
\caption{The problem setup of our proposed \method is different from existing multimodal NMT. A multimodal NMT model takes both text and paired image as the input, while \method takes only sentence in the source language as the usual NMT task. \method synthesizes an image and utilize the internal visual representation to assist translation. }
\label{fig1}
\end{figure}

Visual foundation has been introduced in a novel multimodal Neural Machine Translation (MNMT) task~\citep{specia2016shared,elliott2017findings,barrault2018findings}, which uses bilingual (or multilingual) parallel corpora annotated by images describing sentences' contents (see Figure \ref{fig1}(a)). The superiority of MNMT lies in its ability to use visual information to improve the quality of translation, but its effectiveness largely depends on the availability of data sets, especially the quantity and quality of annotated images. In addition, because the cost of manual image annotation is relatively high, at this stage, MNMT is mostly applied on a small and specific dataset, Multi30K~\cite{elliott2016multi30k}, and is not suitable for large-scale text-only Neural Machine Translation (NMT)~\cite{bahdanau2014neural,vaswani2017attention}. Such limitations hinder the applicability of visual information in NMT.

To address the bottlenecks mentioned above, \citet{zhang2019neural} propose to build a lookup table from an image dataset and then using the search-based method to retrieve pictures that match the source language keywords. 
However, the lookup table is built from Multi30K, which leads to a relatively limited coverage of the pictures, and potentially introduces much irrelevant noise. 
It does not always find the exact image corresponding to the text, or the image may not even exist in the database. 
\citet{elliott2017imagination} present a multitask learning framework to ground visual representation to a shared space. 
Their architecture called ``imagination'' shares an encoder between a primary NMT task and an auxiliary task of ranking the visual features for image retrieval. 
However, neither the image is explicitly generated, nor the visual feature is directly leveraged by the translation decoder, the model simply learns the visual grounded shared encoder. Based on other researchers' earlier exploration, we hypothesize that the potential of vision in conventional text-only NMT has not been fully discovered. Different with \citet{elliott2017imagination} implicit approach, we understand ``imagination'' to be more like ``picturing'', since it is similar to humans who can visually depict figures in the mind from an utterance. Our approach aims to explicitly imagine a ``vague figure'' (see Figure \ref{fig1}(b)) to guide the translation, since \emph{A picture is worth a thousand words}, and imagining the picture of a sentence is the instinctive reaction of a human being who is learning bilingualism.

In this paper, we propose a novel end-to-end machine translation model that is embedded in visual semantics with generative imagination (\method) (see Figure \ref{fig1}(b)). Given a source language sentence, \method first encodes it and transforms the word representations into visual features through an attentive generator, which can effectively capture the semantics of both global and local levels, and the generated visual representations can be considered as semantic-equivalent reconstructions of sentences. A simple yet effective integration module is designed to aggregate the textual and visual modalities. In the final stage, the model learns to generate the target language sentence based on the joint features. To train the model in an end-to-end fashion, we apply a visual realism adversarial loss and a text-image pair-aware adversarial loss, as well as text-semantic reconstruction loss and target language translation loss based on cross-entropy. 


In contrast with most prior MNMT work, our proposed \method model does not require images as input during the inference time but can leverage visual information through imagination, making it an appealing method in low-resource scenario. Moreover, \method is also flexible, accepting external parallel text data or non-parallel image captioning data. We evaluate our Imagination modal on the Multi30K dataset. The experiment results show that our proposed method significantly outperforms the text-only NMT baseline. The analysis demonstrates that imagination help the model complete the missing information in the sentence when we perform degradation masking, and we also see improvements in translation quality by pre-training the model with an external non-parallel image captioning dataset.

To summarize, the paper has the following contributions: 
\begin{enumerate}
    \item We propose generative imagination, a new setup for machine translation assisted by synthesized visual representation, without annotated images as input;
    \item We propose the \method method, which shows advantages over the conventional MNMT model and gains significant improvements over the text-only NMT baseline;
    \item We conduct experiments to verify and analyze how imagination helps the translation.
\end{enumerate}

\section{Related work}
\label{sec:related}
\textbf{MNMT}   As a language shared by people worldwide, visual modality may help machines have a more comprehensive perception of the real world. Multimodal neural machine translation (MNMT) is a novel machine translation task proposed by the machine translation community, which aims to design multimodal translation frameworks using context from the additional visual modality \cite{specia2016shared}. The shared task releases the dataset Multi30K~\cite{elliott2016multi30k}, which is an extended German version of Flickr30K~\cite{young2014image}, then expanded to French and Czech~\cite{elliott2017findings,barrault2018findings}. In the three versions of tasks, scholars have proposed many multimodal machine translation models and methods. \citet{huang2016attention}  encodes word sequences with regional visual objects, while \citet{calixto2017incorporating} study the effects of incorporating global visual features to initialize the encoder/decoder hidden states of RNN. \citet{caglayan2017lium} models the image-text interaction by leveraging element-wise multiplication. \citet{elliott2017imagination} propose a multitask learning framework to ground visual representation to a shared space and learn with the auxiliary triplet alignment task. The common practice is to use convolutional neural networks to extract visual information and then using attention mechanisms to extract visual contexts~\cite{caglayan2016multimodal,calixto2016dcu,libovicky2017attention}. \citet{ive2019distilling}  propose a translate-and-refine approach using two-stage decoder. \citet{calixto2018latent} put forward a latent variable model to capture the multimodal interactions between visual and textual features. \citet{caglayan2019probing} show that visual content is more critical when the textual content is limited or uncertain in MMT. Recently, \citet{yao2020multimodal} propose multimodal self-attention in Transformer to avoid encoding irrelevant information in images, and \citet{yin2020novel} propose a graph-based multimodal fusion encoder to capture various relationships.


\textbf{Text-to-image synthesis}   Traditional Text-to-image (T2I) synthesis mainly uses keywords to search for small image regions, and finally optimizes the entire layout~\cite{zhu2007text}. After generative adversarial networks (GANs)~\cite{goodfellow2014generative} were proposed, scholars have presented a variety of GAN-based T2I models. \citet{reed2016generative} propose DC-GAN and design a direct and straightforward network and a training strategy for T2I generation. \citet{zhang2017stackgan} propose stackGAN, which contains multiple cascaded generators and discriminators, and the higher stage generates better quality pictures. In previous work, scholars only considered global semantics. \citet{xu2018attngan} proposed AttnGAN to apply the attention mechanism to capture fine-grained word-level information. MirrorGAN~\cite{qiao2019mirrorgan} employs a mirror structure, which reversely learns from the inverse task of T2I to further validate whether generated images are consistent with the input texts. The inverse task is also known as image captioning.

\section{ImagiT model}
\label{sec:method}
\begin{figure*}[t]
\centering
\includegraphics[width=0.99\textwidth]{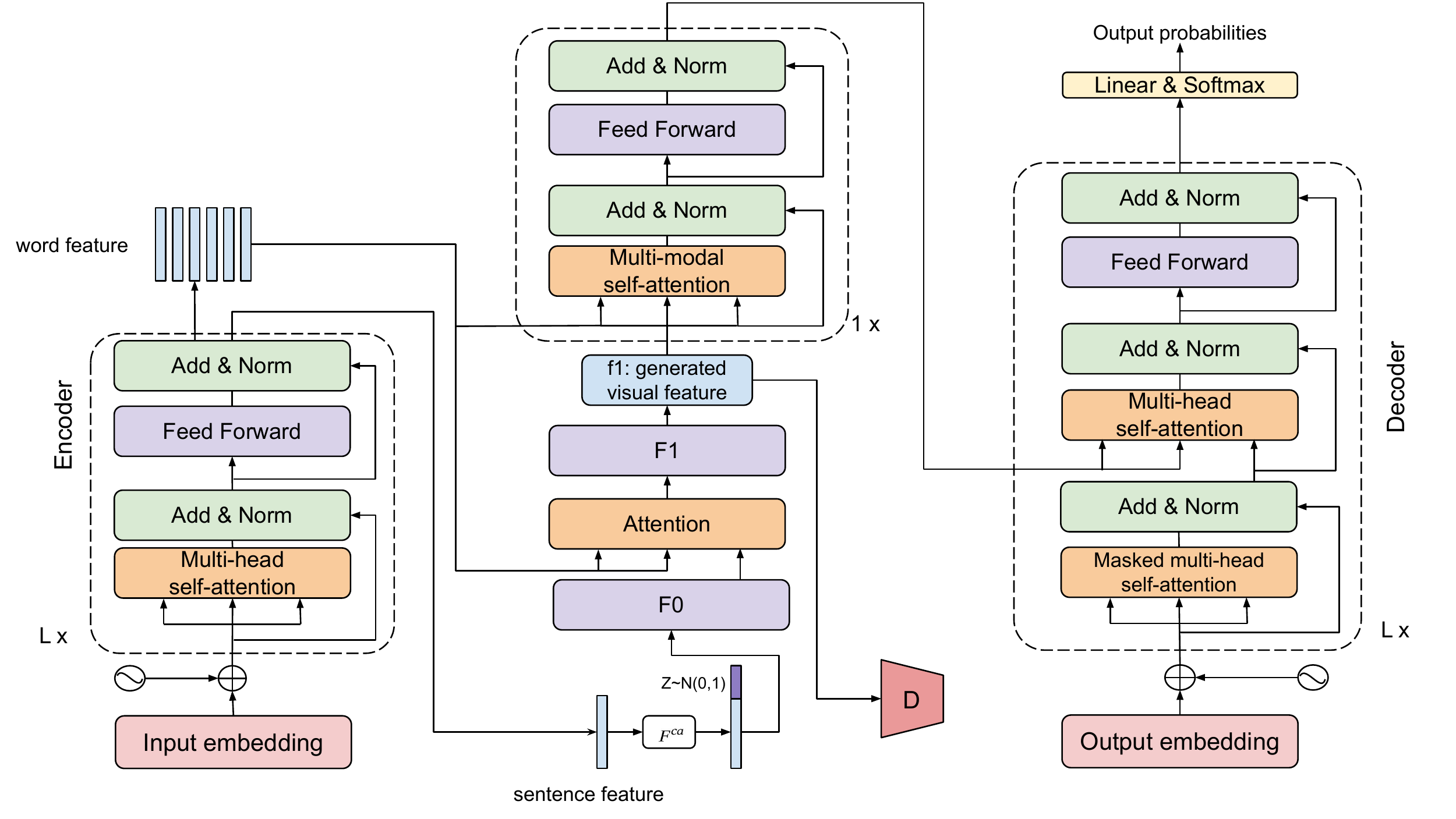} 
\caption{Overview of the framework of the proposed \method. $F_{0}$ and $F_{1}$ are text-to-image converters, sharing similar structures, comprising of perceptron, residual, and unsampling blocks. L$\times$ represents L identical layers. Noting that we only need to obtain the generated visual feature to guide the translation, for the whole pipeline, up-sampling this feature to image is redundant.}
\label{fig2}
\end{figure*}

As shown in Figure\ref{fig2}, \method embodies the encoder-decoder structure for end-to-end machine translation. Between the encoder and the decoder, there is an imagination step to generate semantic-equivalent visual representation. Technically, our model is composed of following modules: source text encoder, generative imagination network, image captioning, multimodal aggregation and decoder for translation. We will elaborate on each of them in the rest of this section.

\subsection{Source text encoder}

\citet{vaswani2017attention} propose the state-of-art Transformer-based machine translation framework, which can be written as follows:

\begin{equation}
    \overline{\textbf{H}}^{l}=\textit{LN}(\textit{Att}^{l}(\textbf{Q}^{l-1},\textbf{K}^{l-1},\textbf{V}^{l-1})+\textbf{H}^{l-1}),
\end{equation}

\begin{equation}
    \textbf{H}^{l}=\textit{LN}(\textit{FFN}^{l}(\overline{\textbf{H}}^{l})+\overline{\textbf{H}}^{l}),
\end{equation}

Where $\textit{Att}^{l}$, $\textit{LN}$, and $\textit{FFN}^{l}$ are the self-attention module, layer normalization, and the feed-forward network for the $l$-th identical layer respectively. The core of the Transformer is the multi-head self-attention, in each attention head, we have:

\begin{equation}
    z_{i}=\sum_{j=1}^{n}\alpha_{ij}(x_{j}W^{V}),
\end{equation}

\begin{equation}
    \alpha_{ij}=softmax(\frac{(x_{i}W^{Q})(x_{j}W^{K})^\top }{\sqrt{d}}).
\end{equation}

$W^{V},W^{Q},W^{K}$ are layer-specific trainable parameter matrices. For the output of final stacked layer, we use $w=\{w_{0},w_{1},...,w_{L-1} \}$, $w\in\mathbb{R}^{d\times L}$ to represent the source word embedding, $L$ is the length of the source sentence. Besides, we add a special token to each source language sentence to obtain the sentence representation $s \in \mathbb{R}^{d}$.

\subsection{Generative imagination network}

Generative Adversarial Network~\cite{goodfellow2014generative} has been applied to synthesis images similar to ground truth~\cite{zhang2017stackgan,xu2018attngan,qiao2019mirrorgan}. We follow the common practice of using the conditioning augmentation~\cite{zhang2017stackgan} to enhance robustness to small perturbations along the conditioning text manifold and improve the diversity of generated samples.\footnote{\citet{zhang2017stackgan} also mentions that the randomness in the Conditioning Augmentation is beneficial for modeling text to image semantic translation as the same sentence usually corresponds to objects with various poses and appearances.} $F^{ca}$ represents the conditioning augmentation function, and $s^{ca}$ represents the enhanced sentence representation.

\begin{equation}
    s^{ca}=F^{ca}(s),
\end{equation}

$\{F_{0}, F_{1}\}$ are two visual feature converters, sharing similar architecture. $F_{0}$ contains a fully connected layer and four deconvolution layers~\cite{Noh2015LearningDN} to obtain image-sized feature vectors. Furthermore, we define $\{f_{0}, f_{1}\}$ are the visual features after two transformations with different resolution. For detailed layer structure and block design, please refer to ~\cite{xu2018attngan}.

\begin{equation}
    f_{0}=F_{0}(z,s^{ca}),
\end{equation}

\begin{equation}
    f_{1}=F_{1}(f_{0},F^{attn}(f_{0},s^{ca})),
\label{equation 7}
\end{equation}

Where $f_{0}\in \mathbb{R}^{M_{0}\times N_{0}}$, $z$ is the noise vector, sampled from the standard normal distribution, and it will be concatenated with $s^{ca}$. Each column of $f_{i}$ is a feature vector of a sub-region of the image, which can also be treat as a pseudo-token. To generate fine-grained details at different subregions of the image by paying attention to the relevant words in the source language, we use image vector in each sub-region to query word vectors by leveraging attention strategy. $F^{attn}$ is an attentive function to obtain word-context feature, then we have:

\begin{small}
\begin{equation}
    F^{attn}(f_{0},s^{ca})=\sum _{l=0}^{L-1} (U_{0}w_{l})(softmax(f_{0}^{T}(U_{0}w_{l})))^\top ,
\end{equation}
\end{small}

Word feature $w_{l}$ is firstly converted into the common semantic space of the visual feature, $U_{0}$ is a perceptron layer. Then it will be multiplied with $f_{0}$ to acquire the attention score. $f_{1}$ is the output of the imagination network, capturing multiple levels (word level and sentence level) of semantic meaning. $f_{1}$ is denoted as the blue block ``generated visual feature'' in Figure\ref{fig2}. It will be utilized directly for target language generation, and it will also be passed to the discriminator for adversarial training. Note that for the whole pipeline, upsampling $f_{1}$ to an image is redundant.

Comparing to T2I synthesis works which use cascaded generators and disjoint discriminators\cite{zhang2017stackgan, xu2018attngan, qiao2019mirrorgan}, we only use one stage to reduce the model size and make our generated visual feature $f_{1}$ focus more on text-mage consistency, but not the realism and authenticity.

\subsection{Image captioning}

Image captioning (I2T) can be regarded as the inverse problem of text-to-image generation, generating the given image's description. If an imagined image is semantic equivalent to the source sentence, then its description should be almost identical to the given text. Thus we leverage the image captioning to translate the imagined visual representation back to the source language\cite{qiao2019mirrorgan}, and this symmetric structure can make the imagined visual feature act like a mirror, effectively enhancing the semantic consistency of the imagined visual feature and precisely reflect the underlying semantics. Following \citet{qiao2019mirrorgan}, we utilize the widely used encoder-decoder image captioning framework\cite{vinyals2015show}, and fix the parameters of the pre-trained image captioning framework when end-to-end training other modules in \method.

\begin{equation}
    p_{t}=Decoder(h_{t-1}),t=0,1,...,L-1,
\end{equation}

\begin{equation}
    \mathcal{L}_{I2T} = -\sum_{t=0}^{L-1} \log p_{t}(T_{t}).
    \label{equation10}
\end{equation}

$p_{t}$ is the predicted probability distribution over the words at $t$-th decoding step, and $T_{t}$ is the $T_{t}$-th entry of the probability vector.

\subsection{Multimodal aggregation}

After obtaining the imagined visual representation, we aggregate two modalities for the translation decoder. Although the vision carries richer information, it also contains irrelevant noise. Comparing to encoding and integrating visual feature directly, a more elegant method is to induce the hidden representation under the guide of image-aware attention and graph perspective of Transformer~\cite{yao2020multimodal}, since each local spatial regions of the image can also be considered as pseudo-tokens, which can be added to the source fully-connected graph. In the multimodal self-attention layer, we add the spatial feature of the generated feature map in the source sentence, that is, the attention query vector is the combination of text and visual embeddings, getting $\Tilde{x}\in\mathbb{R}^{(L+M)\times d}$. Then perform image-aware attention, the key and value vectors are just text embeddings, we have:

\begin{equation}
    c_{i}=\sum_{j=0}^{L-1}\Tilde{\alpha}_{ij}(w_{j}W^{V}),
\end{equation}

\begin{equation}
    \Tilde{\alpha}_{ij}=softmax(\frac{(\Tilde{x}_{i}W^{Q})(w_{j}W^{K})^\top }{\sqrt{d}}).
\end{equation}

\subsection{Objective function}

\begin{figure}[t]
\centering
\includegraphics[width=0.95\columnwidth]{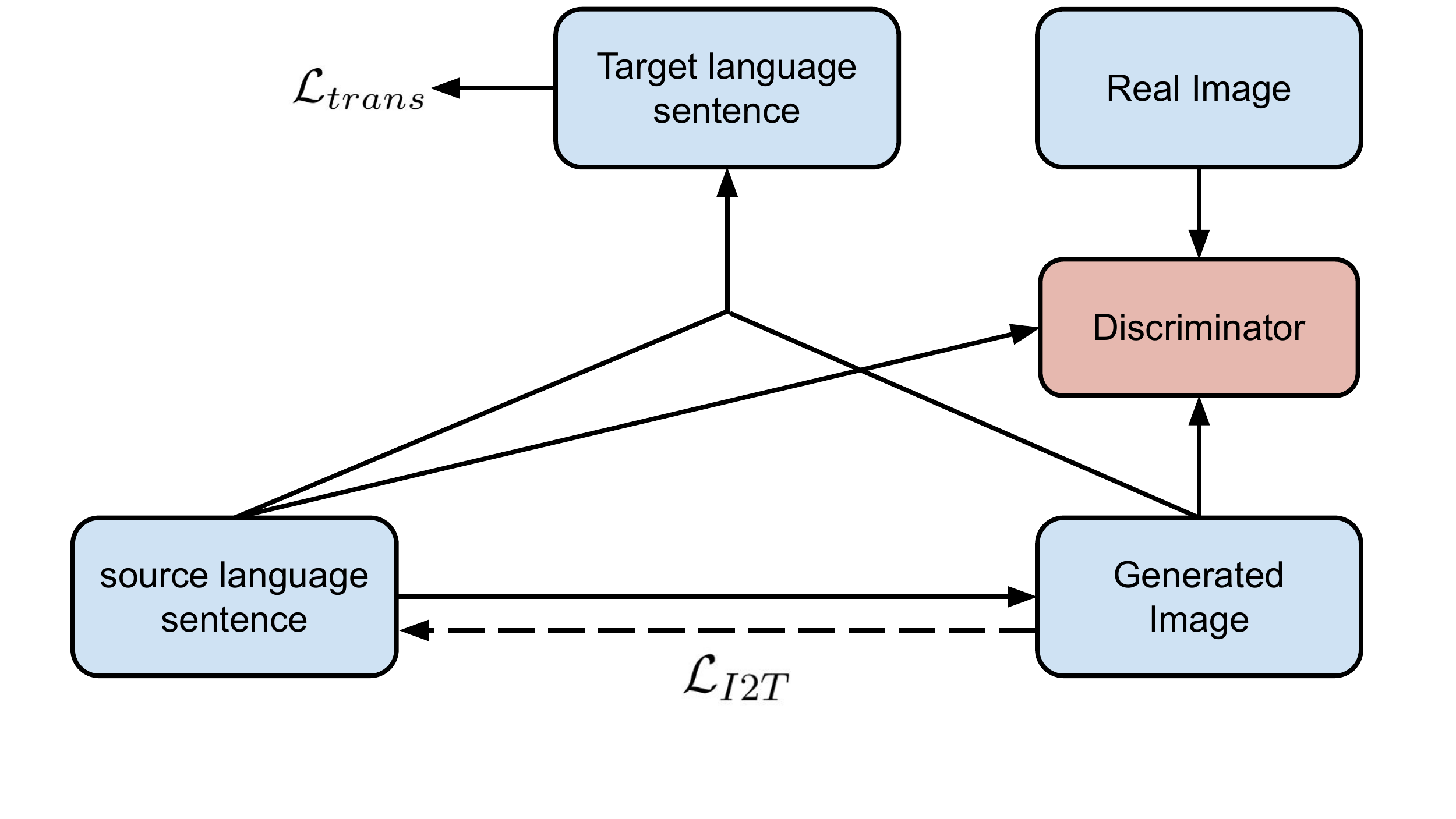} 
\caption{Training objective. The discriminator takes source language sentences, generated images, and real images as input, then computes two adversarial loss: realism loss and text-image paired loss. $\mathcal{L}_{I2T}$ is designed to guarantee the semantic consistency, and $\mathcal{L}_{trans}$ is the core loss function to translate integrated embedding to the target language.}
\label{fig3}
\end{figure}

During the translation phase, similar to equation \ref{equation10}, we have:

\begin{equation}
    \mathcal{L}_{trans} = -\sum_{t} \log p_{t}(T_{t}),
\end{equation}

To train the whole network end-to-end, we leverage adversarial training to alternatively train the generator and the discriminator. Especially, as shown in Figure \ref{fig3}, the discriminator take the imagined visual representation, source language sentence, and the real image as input, and we employ two adversarial losses: a visual realism adversarial loss, and a text-image pair-aware adversarial loss computed by the discriminator~\cite{zhang2017stackgan,xu2018attngan,qiao2019mirrorgan}. 

\begin{equation}
\begin{split}
        \mathcal{L}_{G_{0}}=&-\frac{1}{2}\mathbb{E}_{f_{1}\sim p_{G}}[\log(D(f_{1})]\\ &-\frac{1}{2}\mathbb{E}_{f_{1}\sim p_{G}}[\log(D(f_{1},s)],
\end{split}
\end{equation}

$f_{1}$ is the generated visual feature computed by equation \ref{equation 7} from the model distribution $p_{G}$, $s$ is the global sentence vector. The first term is to distinguish real and fake, ensuring that the generator generates visually realistic images. The second term is to guarantee the semantic consistency between the input text and the generated image. $\mathcal{L}_{G_{0}}$ jointly approximates the unconditional and conditional distributions. The final objective function of the generator is defined as:

\begin{equation}
    \mathcal{L}_{G}=\mathcal{L}_{G_{0}}+\lambda_{1}\mathcal{L}_{I2T}+ \lambda_{2}\mathcal{L}_{trans}.
\label{equ15}
\end{equation}

Accordingly, the discriminator $D$ is trained by minimizing the following loss:

\begin{equation}
\begin{split}
    \mathcal{L}_{D} =&-\frac{1}{2} \mathbb{E}_{I\sim p_{data}}[\log(D(I)]\\
    &-\frac{1}{2} \mathbb{E}_{f_{1}\sim p_{G}}[\log(1-D(f_{1})] \\
    &-\frac{1}{2} \mathbb{E}_{I\sim p_{data}}[\log(D(I,s)]\\
    &-\frac{1}{2} \mathbb{E}_{f_{1}\sim p_{G}}[\log(1-D(f_{1},s)].
\end{split}
\end{equation}

Where $I$ is from the true image distribution $p_{data}$. The first two items are unconditional loss, the latter two are conditional loss.

\begin{table*}[ht]
\centering
\small
\linespread{1.2}
\setlength{\tabcolsep}{1.5mm}{
\begin{tabular}{l|cccc|cccc}
\hline
\multirow{3}{*}{\textbf{Model}} & \multicolumn{4}{c|}{\textbf{En$\Rightarrow$De}} &  \multicolumn{4}{c}{\textbf{En$\Rightarrow$Fr}}\\
\cline{2-9}
& \multicolumn{2}{c}{\textbf{Test2016}}  
& \multicolumn{2}{c|}{\textbf{Test2017}} 
& \multicolumn{2}{c}{\textbf{Test2016}}
& \multicolumn{2}{c}{\textbf{Test2017}}\\
\cline{2-9}
& BLEU & METEOR & BLEU & METEOR & BLEU & METEOR & BLEU & METEOR \\
\hline
\hline
\multicolumn{9}{c}{\emph{Multimodal Neural Machine Translation Systems}} \\
\hline
IMG$_{D}$~\cite{calixto2017incorporating} & 37.3 & 55.1 & N/A & N/A & N/A & N/A & N/A & N/A \\
NMT$_{SRC+IMG}$~\cite{calixto2017doubly}  & 36.5 & 55.0 & N/A & N/A & N/A & N/A & N/A & N/A \\
fusion-conv~\cite{caglayan2017lium} & 37.0 & 57.0 & 29.8 & 51.2 & 53.5 & 70.4 & 51.6 & 68.6 \\
trg-mul~\cite{caglayan2017lium} & 37.8 & \textbf{57.7} & 30.7 & 52.2 & 54.7 & 71.3 & 52.7 & 69.5 \\
VAG-NMT~\cite{zhou2018visual} & N/A & N/A & 31.6 & 52.2 & N/A& N/A& \textbf{53.8} & \textbf{70.3} \\
Transformer+Att~\cite{ive2019distilling} & 38.0 & 55.6 & N/A & N/A & 59.8 & \textbf{74.4} & N/A & N/A \\
Multimodal~\cite{yao2020multimodal} & \textbf{38.7} & 55.7 & N/A & N/A & N/A & N/A & N/A & N/A\\
\method + ground truth & 38.6 & 55.7 & \textbf{32.4} & \textbf{52.5} & \textbf{59.9} & 74.3 & 52.8 & 68.6\\
\hline
\multicolumn{9}{c}{\emph{Text-only Neural Machine Translation Systems}} \\
\hline
Transformer~\cite{vaswani2017attention} & 37.6 & 55.3 & 31.7 & 52.1 & 59.0 & 73.6 & 51.9 & 68.3 \\
Multitask~\cite{elliott2017imagination} & 36.8 & 55.8 & N/A & N/A & N/A & N/A & N/A & N/A \\
VMMT$_{F}$~\cite{calixto2018latent} & 37.6 & 56.0 & N/A & N/A & N/A & N/A & N/A & N/A\\
Lookup table~\cite{zhang2019neural} & 36.9 & N/A & 28.6 & N/A & 57.5 & N/A & 48.5 & N/A\\
\method & 38.5 & 55.7 & 32.1 & 52.4 & 59.7 & 74.0 & 52.4 & 68.3 \\
\hline
\end{tabular}}
\caption{Main result from the Test2016, Test2017 for the En$\Rightarrow$De and En$\Rightarrow$Fr MNMT task. The first category (Multimodal Neural Machine Translation Systems) collects the existing MNMT systems, which take both source sentences and paired images as input. The second category illustrates the systems that do not require images as input. Since our method falls into the second group, the baselines are the text-only 
Transformer~\cite{vaswani2017attention} and the aforementioned works~\cite{zhang2019neural,elliott2017imagination}.}
\label{Table_En2DeMainResults}
\end{table*}

\section{Experiments}
\label{sec:exp}
\subsection{Datasets}

We evaluate our proposed \method model on two datasets, Multi30K~\cite{elliott2016multi30k} and Ambiguous COCO~\cite{elliott2017findings}. To show its ability to train with external out-of-domain datasets, we adopt MS COCO~\cite{lin2014microsoft} in the next analyzing section.

Multi30K is the largest existing human-labeled collection for MNMT, containing $31K$ images and consisting of two multilingual expansions of the original Flickr30K\cite{young2014image} dataset. The first expansion has five English descriptions and five German descriptions, and they are independent of each other. The second expansion has one of its English description manually translated to German by a professional translator, then expanded to French and Czech in the following shared task~\cite{elliott2017findings,barrault2018findings}. We only apply the second expansion in our experiments, which has $29,000$ instances for training, $1,014$ for development, and $1,000$ for evaluation. We present our results on English-German (En-De) English-French (En-Fr) Test2016 and Test2017.

Ambiguous COCO is a small evaluation dataset collected in the WMT2017 multimodal machine translation challenge~\cite{elliott2017findings}, which collected and translated a set of image descriptions that potentially contain ambiguous verbs. It contains $461$ images from the MS COCO\cite{lin2014microsoft} for $56$ ambiguous vers in total. 

MS COCO is the widely used non-parallel text-image paired dataset in T2I and I2T generation. It contains $82,783$ training images and $40,504$ validation images with $91$ different object types, and each image has 5 English descriptions.

\subsection{Settings}

\begin{table*}[ht]
\centering
\small
\linespread{1.2}
\setlength{\tabcolsep}{1.5mm}{
\begin{tabular}{l|cc|cc}
\hline
\multirow{3}{*}{\textbf{Model}} & \multicolumn{2}{c|}{\textbf{En$\Rightarrow$De}} & \multicolumn{2}{c}{\textbf{En$\Rightarrow$Fr}}\\
\cline{2-5}
& \multicolumn{2}{c|}{\textbf{Ambiguous COCO}}
& \multicolumn{2}{c}{\textbf{Ambiguous COCO}}\\
\cline{2-5}
& BLEU & METEOR & BLEU & METEOR\\
\hline
\hline
\multicolumn{5}{c}{\emph{Multimodal Neural Machine Translation Systems}} \\
\hline
fusion-conv~\cite{caglayan2017lium} & 25.1 & 46.0 & 43.2 & 63.1 \\
trg-mul ~\cite{caglayan2017lium} & 26.4 & 47.4 & 43.5 & 63.2 \\
VAG-NMT ~\cite{zhou2018visual} & 28.3 & 48.0 & 45.0 & 64.7 \\
\method + ground truth & \textbf{28.8} & \textbf{48.9} & \textbf{45.3} & \textbf{65.1}\\
\hline
\multicolumn{5}{c}{\emph{Text-only Neural Machine Translation Systems}} \\
\hline
Transformer baseline ~\cite{vaswani2017attention} & 27.9 & 47.8 & 44.9 & 64.2\\
\method & 28.7 & 48.8 & \textbf{45.3} & 65.0\\
\hline
\end{tabular}}
\caption{Experimental results on the Ambiguous COCO En$\Rightarrow$De and En$\Rightarrow$Fr translation task.}
\label{MSCOCO}
\end{table*}

Our baseline is the conventional text-only Transformer~\cite{vaswani2017attention}. Specifically, each encoder-decoder has a 6-layer stacked Transformer network, eight heads, 512 hidden units, and the inner feed-forward layer filter size is set to 2048. The dropout is set to $p=0.1$, and we use Adam optimizer~\cite{kingma2014adam} to tune the parameter. The learning rate increases linearly for the warmup strategy with $8,000$ steps and decreases with the step number's inverse square root. We train the model up to $10,000$ steps, the early-stop strategy is adopted. We use the same setting as \citet{vaswani2017attention}. We use the metrics BLEU~\cite{papineni-etal-2002-bleu} and METEOR~\cite{denkowski-lavie-2014-meteor}to evaluate the translation quality.

For the imagination network, the noise vector's dimension is $100$, and the generated visual feature is $128\times128$. The upsampling and residual block in visual feature transformers consist of $3\times3$ stride 1 convolution, batch normalization, and ReLU activation. The training is early-stopped if the dev set BLEU score do not improve for $10$ epochs, since the translation is the core task. The batch size is $64$, and the learning rate is initialized to be $2e^{-4}$ and decayed to half of its previous value every $100$ epochs. A similar learning schedule is adopted in \citet{zhang2017stackgan}. The margin size $\gamma$ is set to $0.1$, the balance weight $\lambda_{1}=20$, $\lambda_{2}=40$. 

\subsection{Results}

Table \ref{Table_En2DeMainResults} illustrates the results for the En-De Test2016, En-De Test2017, En-Fr Test2016 and En-Fr Test2017 tasks. Our text-only Transformer baseline~\cite{vaswani2017attention} has similar results compared to most prior MNMT works, which is consistent with the previous findings~\cite{caglayan2019probing}, that is, textual modality is good enough to translate for Multi30K dataset. This finding helps to explain that it is already tricky for a MNMT model to ground visual modality even with the presence of annotated images. However, Our \method gains improvements over the text-only Transformer baseline on four evaluation datasets, demonstrating that our model can effectively embed the visual semantics during the training time and guide the translation through imagination with the absence of annotated images during the inference time. We assume much of the performance improvement is due to \method's strong ability to capture the interaction between text and image, generate semantic-consistent visual representations, and incorporate information from visual modality properly. 

We also observe that our approach surpasses the results of most MNMT systems by a noticeable margin in terms of BLEU score and METEOR score on four evaluation datasets. Our \method is also competitive with \method + ground truth, which is our translation decoder taking ground truth visual representations instead of imagined ones, and can be regarded as the upper boundary of imagiT. This proves imaginative ability of \method.

Table \ref{MSCOCO} shows results for the En-De En-Fr Ambiguous COCO. For Ambiguous COCO, which was purposely curated such that verbs have ambiguous meaning, demands more visual contribution for guiding the translation and selecting correct words. Our \method benefits from visual imagination and substantially outperforms previous works on ambiguous COCO. and even gets the same performance as \method + ground truth (45.3 BLEU).

\subsection{Ablation studies}

The hyper-parameter $\lambda_{1}$ in equation \ref{equ15} is important. When $\lambda_{1}=0$, there is no image captioning component, the BLEU score drops from 38.5 to 37.9, while this variant still outperforms the Transformer baseline. This indicates the effectiveness of image captioning module, since it will potentially prevent visual-textual mismatching, thus helps generator achieve better performance. When $\lambda_{1}$ increases from $5$ to $20$, the BLEU and METEOR increase accordingly. Whereas $\lambda_{1}$ is set to equal to $\lambda_{2}$, the BLEU score falls to 38.3. That's reasonable because $\lambda_{2}\mathcal{L}_{trans}$ is the main task of the whole model.

\begin{table}[h]
\centering
\begin{tabular}{l|cc}
\hline
 Evaluation metric             & BLEU   & METEOR     \\ \hline
 ImagiT, $\lambda_{1}=0$ & 37.9 & 55.3 \\ \hline
  ImagiT, $\lambda_{1}=5$ & 38.2 & 55.5 \\ \hline
   ImagiT, $\lambda_{1}=10$ & 38.4 & 55.7 \\ \hline
    ImagiT, $\lambda_{1}=20$ & 38.5 & 55.7 \\ \hline
     ImagiT, $\lambda_{1}=40$ & 38.3 & 55.6 \\ \hline
\end{tabular}
\caption{Ablation studies of ImagiT with different weight settings}
\label{table2}
\end{table}

\section{Analysis}
\label{sec:ana}
\subsection{Can \method generate visual grounded representations?}

Since the proposed model does not require images as input, one may ask how it uses visual information and where the information comes? We claim that \method has already been embedded with visual semantics during the training phase, and in this section, we validate that \method is able to generate visual grounded representation by performing the image retrieval task.

For each source sentence, we generate the intermediate visual representation. Furthermore, we query the ground truth image features for each generated representation to find the closest image vectors around it based on the cosine similarity. Then we can measure the $R@K$ score, which computes the recall rate of the matched image in the top K nearest neighborhoods.

\begin{table}[h]
\centering
\begin{tabular}{l|ccc}
\hline
              & $R@1$     & $R@5$ &  $R@10$   \\ \hline
\method on Multi30K & 64.7  & 88.7 & 94.2 \\ \hline
\method on MS COCO & 64.3  &  89.5   & 94.7 \\
\hline
\end{tabular}
\caption{Image retrieval task. We evaluate on Multi30K and MS COCO.}
\label{retrieval}
\end{table}

Some previous studies on VSE perform sentence-to-image retrieval and image-to-sentence retrieval, but their results can not be directly compared with ours, since we are performing image-to-image retrieval in practical. However, from Table \ref{retrieval}, especially for $R@10$, the results demonstrate that our generated representation has excellent quality of shared semantics and have been grounded with visual semantic-consistency. 

\subsection{How does the imagination help the translation?}

Although we have validated the effectiveness of \method on three widely used MNMT evaluation datasets. A natural question to ask is that how does the imagination guide the translation, and to which extent? When human beings confronting with complicate sentences and obscure words, we often resort to mind-picturing and mental visualization to assist us to auto-complete and fill the whole imagination. Thus we hypothesis that imagination could help recover and retrieve the missing and implicate textual information.

Inspired by \citet{ive2019distilling,caglayan2019probing}, we apply degradation strategy to the input source language, and feed to the trained Transformer baseline, MNMT baseline, and \method respectively, to validate if our proposed approach could recover the missing information and obtain better performance. And we conduct the analysing experiments on En-De Test2016 evaluation set.

\textbf{Color deprivation} is to mask the source tokens that refers to colors, and replace them with a special token [M]. Under this circumstance, text-only NMT model have to rely on source-side contextual information and biases, while for MNMT model, it can directly utilize the paired color-related information-rich images. But for \method, the model will turn to imagination and visualization.

\begin{table}[!h]
\centering
\begin{tabular}{l|cc}
\hline
 Model             & $S$     & $\overline{S}$    \\ \hline
text-only Transformer & 37.6  & 36.3 \\ \hline
MNMT        & 38.2 & 37.7 \\ \hline
\method       & 38.4 & 37.9 \\
\hline
\end{tabular}
\caption{Color deprivation. $s$ represents the original source sentence, while $\overline{s}$ is the degraded sentence.}
\label{color deprivation}
\end{table}

Table ~\ref{color deprivation} demonstrates the results of color deprivation. We implement a simple transformer-based MNMT baseline model using the multimodal self-attention approach~\cite{yao2020multimodal}. Thus the illustrated three models in Table \ref{color deprivation} can be compared directly. We can observe that the BLEU score of text-only NMT decreases 1.3, whereas MNMT and \method system only decreases 0.5. This result corroborates that our \method has a similar ability to recover color compared to MNMT, but our \method achieves the same effect through its own efforts, i.e., imagination. One possible explanation is that \method could learn the correlation and co-occurrence of the color and specific entities during the training phase, thus imagiT could infer the color from the context and recover it by visualization.

\textbf{Visually depictable entity masking}. \citet{plummer2015flickr30k} extend Flickr30K with cereference chains to tag mentions of visually depictable entities. Similar to color deprivation, we randomly replace $0\%,15\%,30\%,45\%,60\%$ visually depictable entities with a special token [M].

\begin{figure}[ht]
\centering
\includegraphics[width=0.85\columnwidth]{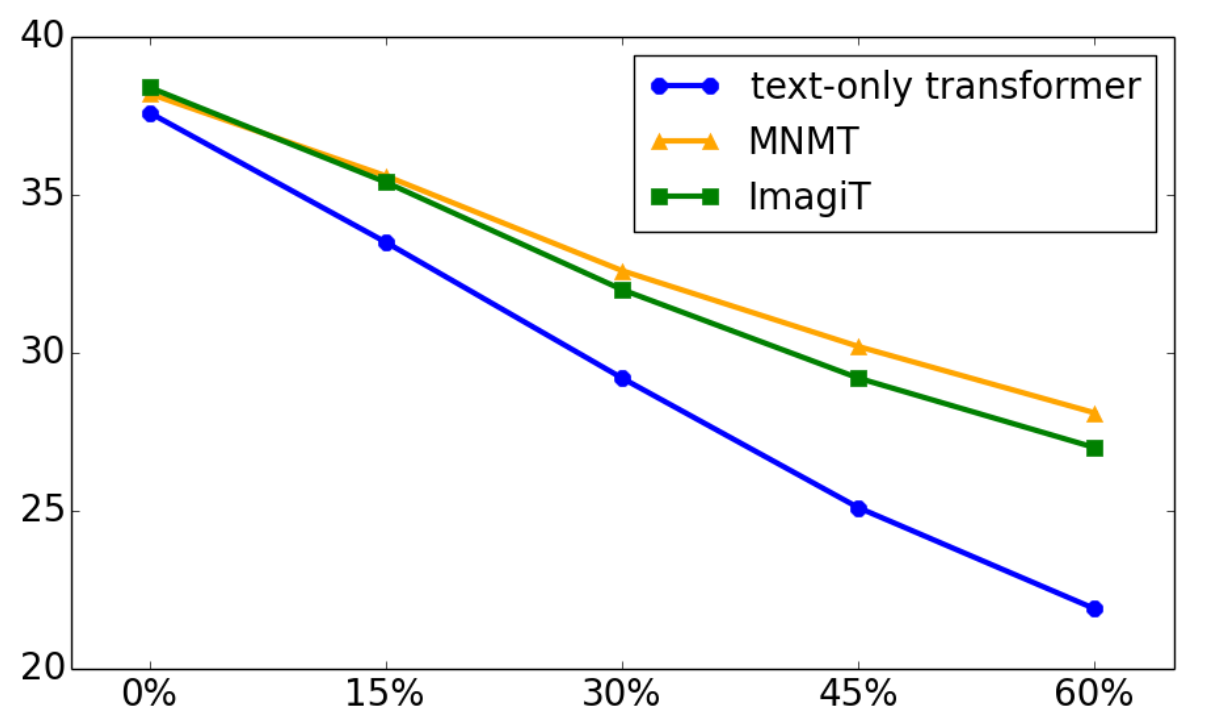} 
\caption{Visually depictable entity masking. From top to bottom is MNMT, \method, text-only transformer.}
\label{fig4}
\end{figure}

Figure \ref{fig4} is the result of visually depictable entity masking. We observe a large BLEU score drop of text-only Transformer baseline with the increasing of masking proportion, while MNMT and \method are relatively smaller. This result demonstrates that our \method model can much more effectively infer and imagine missing entities compared to text-only Transformer, and have comparable capability over the MNMT model.

\subsection{Will better imagination with external data render better translation?}

Our \method model also accepts external parallel text data or non-parallel image captioning data, and we can easily modify the objective function to train with out-of-domain non-triple data. To train with text-image paired image captioning data, we can pre-train our imagination model by ignoring $\mathcal{L}_{trans}$ term~\cite{yang2020towards}. In other words, the T2I synthesis module can be solely trained with MS COCO dataset. We randomly split MS COCO in half, and use COCO$_{half}$ and COCO$_{full}$ to pre-train \method. The MS COCO is processed using the same pipeline as in \citet{zhang2017stackgan}. Furthermore, the training setting of COCO$_{half}$ and COCO$_{full}$ are the same with batch size $64$ and maximum epoch $600$. The results are:

\begin{table}[!h]
\centering
\begin{tabular}{l|cc}
\hline
              & BLEU     & METEOR    \\ \hline
\method & 38.4  & 55.7 \\ \hline
\method + COCO$_{half}$& 38.6 & 56.3 \\ \hline
\method + COCO$_{full}$& 38.7 & 56.7 \\
\hline
\end{tabular}
\caption{Translation results when using out-of-domain non-parallel image captioning data.}
\label{table3}
\end{table}

As is shown in Table \ref{table3}, our \method model pre-trained with half MS COCO gain 0.6 METEOR increase, and the improvement becomes more apparent when training with the whole MS COCO. We can contemplate that large-scale external data may further improve the performance of \method, and we have not utilized parallel text data (e.g., WMT), even image-only and monolingual text data can also be adopted to enhance the model capability, and we leave this for future work.

\section{Conclusion}
\label{sec:conclusion}
This work presents generative imagination-based machine translation model (\method), which can effectively capture the source semantics and generate semantic-consistent visual representations for imagination-guided translation. Without annotated images as input, out model gains significant improvements over text-only NMT baselines and is comparable with the SOTA MNMT model. We analyze how imagination elevates machine translation and show improvement using external image captioning data. Further work may center around introducing more parallel and non-parallel, text, and image data for different training schemes. 

\section{Broader Impact}
\label{sec:broader}
This work brings together text-to-image synthesis, image captioning, and neural machine translation (NMT) for an adversarial learning setup, advancing the traditional NMT to utilize visual information. For multimodal neural machine translation (MNMT), which possesses annotated images and can gain better performance, manual image annotation is costly, so MNMT is only applied on a small and specific dataset. This work tries to extend the applicability of MNMT techniques and visual information in NMT by imagining a semantic equivalent picture and making it appropriately utilized by visual-guided decoder. Compared to the previous multimodal machine translation approaches, this technique takes only sentences in the source languages as the usual machine translation task, making it an appealing method in low-resource scenarios. However, the goal is still far from being achieved, and more efforts from the community are needed for us to get there. One pitfall of our proposed model is that trained \method is not applicable to larger-scale text-only NMT tasks, such as WMT'14, which is mainly related to economies and politics, since those texts are not easy to be visualized, containing fewer objects and visually depictable entities. We advise practitioners who apply visual information in large-scale text-to-text translation to be aware of this issue. In addition, the effectiveness of MNMT model largely depends on the quantity and quality of annotated images, likewise, our model performance also depends on the quality of generated visual representations. We will need to carefully study how the model balance the contribution of different modality and response to ambiguity and bias to avoid undesired behaviors of the learned models.


\bibliographystyle{acl_natbib}

\end{document}